\documentclass[sigplan, 10pt, screen, authorversion, nonacm]{acmart}

\usepackage{tikz}
\usepackage{amsmath}

\usepackage{graphicx}
\usepackage{subcaption}
\usepackage{ulem}
\usepackage{cases}
\usepackage{multirow}
\usepackage{colortbl}
\usepackage{booktabs}
\usepackage{hhline}
\usepackage{makecell} 
\usepackage{diagbox}
\usepackage{pifont}
\usepackage{breqn}

\usepackage{filecontents}

\begin{document}

\date{}

\title{Fate: \underline{Fa}s\underline{t} \underline{E}dge Inference of Mixture-of-Experts Models via Cross-Layer Gate}

\author{Zhiyuan Fang}
\affiliation{
  \institution{Sun Yat-sen University}
  \city{Zhuhai}
  \country{China}
}
\email{fangzhy27@mail2.sysu.edu.cn}

\author{Zicong Hong}
\affiliation{
  \institution{Hong Kong University of Science and Technology}
  \city{Hong Kong}
  \country{China}
}
\email{ziconghong@gmail.com}

\author{Yuegui Huang}
\affiliation{
  \institution{Sun Yat-sen University}
  \city{Guangzhou}
  \country{China}
}
\email{huangyg35@mail3.sysu.edu.cn}

\author{Yufeng Lyu}
\affiliation{
 \institution{Huawei Technologies Co. Ltd}
 \city{Shenzhen}
 \country{China}
}
\email{lvyufeng1@huawei.com}

\author{Wuhui Chen}
\authornotemark[1]
\affiliation{
  \institution{Sun Yat-sen University}
  \city{Zhuhai}
  \country{China}
}
\affiliation{
  \institution{Peng Cheng Laboratory}
  \city{Shenzhen}
  \country{China}
}
\email{chenwuh@mail.sysu.edu.cn}

\author{Yue Yu}
\affiliation{
  \institution{Peng Cheng Laboratory}
  \city{Shenzhen}
  \country{China}
}
\email{yuy@pcl.ac.cn}

\author{Fan Yu}
\affiliation{
  \institution{Huawei Technologies Co. Ltd}
  \city{Shenzhen}
  \country{China}
}
\email{fan.yu@huawei.com}

\author{Zibin Zheng}
\affiliation{
  \institution{Sun Yat-sen University}
  \city{Zhuhai}
  \country{China}
}
\email{zhzibin@mail.sysu.edu.cn}

\begin{abstract}
Large Language Models (LLMs) have demonstrated impressive performance across various tasks, and their application in edge scenarios has attracted significant attention. However, sparse-activated Mixture-of-Experts (MoE) models, which are well suited for edge scenarios, have received relatively little attention due to their high memory demands. Offload-based methods have been proposed to address this challenge, but they face difficulties with expert prediction. Inaccurate expert predictions can result in prolonged inference delays.
To promote the application of MoE models in edge scenarios, we propose Fate, an offloading system designed for MoE models to enable efficient inference in resource-constrained environments. The key insight behind Fate is that gate inputs from adjacent layers can be effectively used for expert prefetching, achieving high prediction accuracy without additional GPU overhead. Furthermore, Fate employs a shallow-favoring expert caching strategy that increases the expert hit rate to 99\%. Additionally, Fate integrates tailored quantization strategies for cache optimization and IO efficiency.
Experimental results show that, compared to Load on Demand and Expert Activation Path-based method, Fate achieves up to 4.5× and 1.9× speedups in prefill speed and up to 4.1× and 2.2× speedups in decoding speed, respectively, while maintaining inference quality. Moreover, Fate's performance improvements are scalable across different memory budgets.
\end{abstract}

\maketitle
\section{Introduction}
Large Language Models (LLMs) have demonstrated extraordinary capabilities in various domains~\cite{nijkampcodegen, xuparadigm, wang2023element}. Currently, the most powerful models~\cite{achiam2023gpt, team2024gemini, Anthropic}, supporting widely used applications, are running on large-scale clusters comprising cutting-edge GPUs (e.g., NVIDIA H100~\cite{NVIDIA_H100}) to provide high-quality services to users worldwide. Recently, interest has shifted to deploying LLMs in edge scenarios to achieve better privacy protection, personalization, and reduced inference latency, such as autonomous driving~\cite{hu2023planning, tian2024drivevlm}, robotics~\cite{zeng2023large, Figure}, and IoT devices~\cite{xu2024general}. In addition, many exciting applications of on-device LLMs have emerged~\cite{yao2024minicpm, Qualcomm}, significantly improving the user experience.

\begin{figure}
  \centering
  \includegraphics[width=\linewidth]{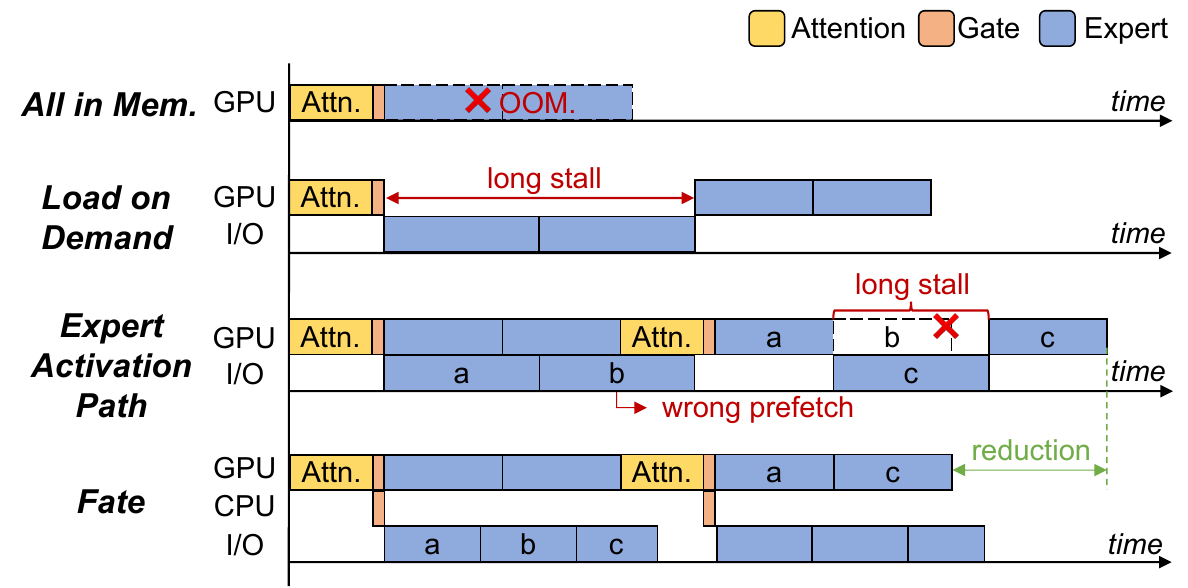}
  \caption{Comparison of three kinds of pipeline.}
  \label{figure:pipeline}
\end{figure}

Current research and applications of on-device LLMs focus primarily on dense models~\cite{qwen2, dubey2024llama, xu2024empowering}. Benefiting from the accelerated growth in the knowledge density of language models, 3-billion-parameter models are well suited for providing high-quality inference services within the resource constraints of edge devices~\cite{qwen2}. Meanwhile, Mixture of Experts (MoE) models~\cite{fedus2022switch, rajbhandari2022deepspeed}, which have gained popularity for improving computational efficiency, are rarely used on edge devices due to their enormous parameter sizes. However, MoE models are actually well suited for edge scenarios because of their sparse activation characteristics. They can achieve superior performance with low resource requirements by activating only a small subset of parameters. For example, Qwen1.5-MoE~\cite{qwen_moe} activates only 2.7 billion parameters during inference, providing better performance and faster inference speeds than many dense 7-billion-parameter models.

The memory bottleneck is the main challenge in applying MoE models to the edge. MoE replaces the MLP layers in Transformer architectures with MoE layers, which consist of a top-k gating network and multiple experts. During each inference step, the gate selects only the top-k experts to participate in the computation, reducing the computational cost. However, the multi-expert architecture dramatically increases the memory requirements, making it infeasible to load all parameters into memory. For instance, Qwen1.5-MoE, which contains 14.3 billion parameters, requires nearly 30 GB of memory for inference. In contrast, NVIDIA RTX 1080Ti GPU, commonly used at the edge, provides high computational power, but is limited to 11 GB of memory.

Offloading is a direct and effective way to alleviate memory bottlenecks. It releases weights that are not currently needed from memory, significantly reducing memory requirements. Since experts make up the majority of parameters in MoE models, strategies designed for MoE typically focus on offloading experts. However, loading experts on demand severely impacts inference latency. For example, when using a top-2 gate, the pipelines of different loading strategies are shown in~\autoref{figure:pipeline}. Load on Demand, a naive approach, introduces significant pipeline bubbles that slow down the inference speed, an unacceptable drawback for latency-sensitive edge scenarios. This is due to the fact that expert activation is highly dependent on the gating results, making it difficult to determine which experts to load in advance. Overcoming this challenge is central to improving the efficiency of MoE inference with offloading.

To address the challenges, existing studies propose various prefetching strategies, which can be broadly categorized into two approaches: expert activation path-based methods (EAP) and small-network-based methods. The path-based approaches, such as Lina~\cite{li2023accelerating} and MoE-Infinity~\cite{xue2024moe}, leverage the correlation (frequency relationship) between the experts selected by the tokens in different layers. During the inference, the selection of a token in the previous $i$ layers is recorded in expert activation path, and then the expert selection in the next layer is predicted based on the correlation. However, these methods suffer from low prediction accuracy.
On the other hand, the small-network-based approach trains an additional lightweight network to predict or replace the original gates, achieving significantly higher accuracy. For instance, SiDA~\cite{du2024sida} and Pre-Gated MoE~\cite{hwang2024pre} utilize a hashing network and a redesigned gate, respectively, to guide expert selection. Nevertheless, this approach introduces additional training costs, and limited portability, and often requires fine-tuning the model to maintain performance.

To enable fast inference of MoE models in edge scenarios, several challenges need to be addressed. First, a high-accuracy and low-cost expert prefetching strategy is essential. Misfetches resulting in temporary I/O can significantly degrade inference latency. Secondly, overlapping computation with I/O remains a challenge. 
Due to the physical limitation that computation is faster than I/O, the GPU often waits for the expert transfer to complete, resulting in a waste of computational resources.
Additionally, most existing approaches overlook the differences between the prefill and decoding stages of MoE inference. The prefill phase of MoE models is not only computationally intensive but also I/O intensive. Processing all of the tokens in the input often results in the activation of a large number of experts, creating far greater I/O and computational pressure than the decoding phase. 

To address this challenge, we propose Fate, an MoE inference system designed to enable efficient MoE inference in edge scenarios, thereby facilitating the deployment of MoE models on resource-constrained devices. The key observation underlying Fate is that the input from the previous gate can be effectively utilized by the next gate for expert selection, enabling highly accurate expert prefetching without incurring additional GPU overhead. In addition, Fate identifies the lower prefetch accuracy in shallow layers and introduces a shallow-favoring expert caching strategy to improve expert cache hit rates. Finally, Fate leverages quantization techniques to increase the number of cached experts and accelerate I/O operations, further improving overall inference efficiency.

In summary, this paper makes the following contributions:

\begin{itemize}
\item We propose a cross-layer expert prefetch, which achieves a prefetch accuracy of 97. 15\%. Furthermore, we introduce the shallow-favoring expert cache, which increases the expert hit rate to 99.08\%. To reduce I/O overhead while maintaining inference accuracy, we also present a popularity-aware hybrid quantization strategy.
\item We implement the above strategies in Fate, an MoE-oriented inference system, which enables fast inference of MoE models in edge scenarios.
\item To evaluate the performance of Fate, we conducted experiments to measure its impact on inference efficiency. The results show that Fate significantly improves the inference efficiency of MoE models in edge scenarios, achieving up to 4.1× and 2.2× decoding speedup compared to Load on Demand and EAP, while maintaining high inference quality.
\end{itemize}

\section{Background}

\subsection{A Primer on MoE Model}

\begin{figure}[]
    \centering
    \begin{subfigure}[b]{0.48\linewidth}
        \centering
        \includegraphics[width=\linewidth]{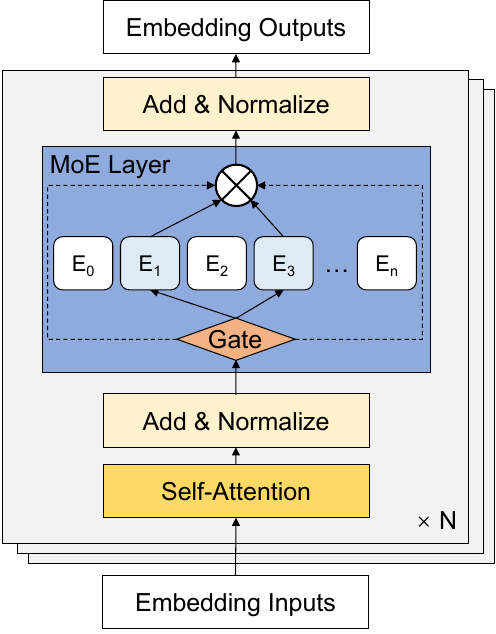} 
        \caption{Conventional top-k MoE model.}
        \label{fig:top-k}
    \end{subfigure}
    \hfill
    \begin{subfigure}[b]{0.48\linewidth}
        \centering
        \includegraphics[width=\linewidth]{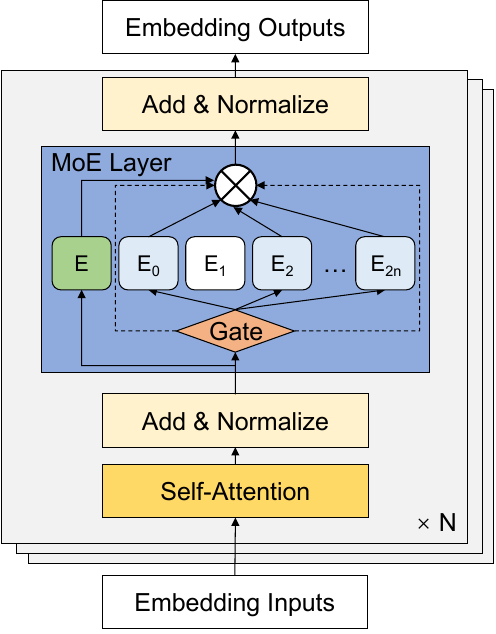} 
        \caption{MoE model with shared and fine-grained experts.}
        \label{fig:fine-grained}
    \end{subfigure}

    \caption{Two mainstream structural designs of MoE model.}
    \label{figure:MoE}
\end{figure}

To further scale the model size without increasing the training and inference costs, some works have introduced the sparsely activated MoE architecture into LLM~\cite{fedus2022switch, shazeer2017outrageously}. Due to its ability to maintain high computational efficiency at relatively low cost, MoE has rapidly become one of the mainstream architectures for LLMs. Representative examples include GPT-4~\cite{achiam2023gpt}, Gemini 1.5~\cite{team2024gemini}, and DeepSeek-V2~\cite{deepseekv2}, etc. Currently, MoE models can be categorized into two main structural designs, as illustrated in~\autoref{figure:MoE}.


In the conventional architecture of MoE models, shown in~\autoref{fig:top-k}, the FFN (Feed-Forward Network) layer in the Transformer is replaced by an MoE layer consisting of a gating network and multiple FFNs (experts).
The gating mechanism is the core concept of MoE architectures. It is typically implemented as a softmax function that activates the top-k experts based on a probability distribution derived from the input. The original intention of introducing multiple experts was to enable different experts to specialize in learning and processing information from different domains.
However, studies~\cite{jiang2024mixtral} have shown that experts tend to focus more on syntactic structures rather than domain-specific content. 

To further enhance expert specialization, DeepSeekMoE~\cite{dai2024deepseekmoe} introduces a shared and fine-grained expert architecture, as illustrated in~\autoref{fig:fine-grained}. Under the constraint of maintaining the same total number of expert parameters, each expert is divided into smaller, fine-grained experts along the intermediate dimension. More fine-grained experts are activated to keep the computational cost stable. This subdivision improves the specialization level of each expert, allowing for more precise knowledge distribution and flexible combinations of experts. Additionally, a subset of experts is isolated as shared experts, which are always activated to capture general knowledge, reduce parameter redundancy across routed experts, increase parameter efficiency, and ensure that each remaining expert focuses on its specific task.
In addition to DeepSeekMoE, other models such as DBRX~\cite{mosaic2024introducing}, Qwen1.5-MoE~\cite{qwen_moe}, and OLMoE~\cite{muennighoff2024olmoe} also demonstrated the effectiveness of using fine-grained experts.

\subsection{Inference of MoE models}

Similar to traditional dense LLMs, the inference process of MoE models follows an autoregressive paradigm, where tokens are generated sequentially, with each token depending on previously generated ones. This process can be broadly divided into two key stages: prefill and decoding.

In the prefill stage, the model processes all tokens of the input sequence in parallel. During this phase, MoE models exhibit higher computational parallelism because tokens can be fed into the model simultaneously, allowing multiple experts to be activated. However, when experts are distributed across heterogeneous hardware, activating a large number of experts at the same time means high IO costs for loading experts. This has a direct impact on TTFT (Time to First Token).

Unlike the parallel computation in the prefill stage, the decoding stage operates sequentially, where the model generates one token at a time, using previously generated tokens as context. Each token is allocated to the top-k experts for computation at each layer through a gating mechanism. The performance of the decoding stage is typically measured by TPOT (Tokens Per Output Time), which represents the number of tokens that can be generated per unit time.

Although the MoE models can achieve high computational efficiency at a low cost, the large number of parameters makes it difficult to deploy them in edge scenarios, and the fine-grained expert design makes expert prediction  more difficult.

\section{Motivation}


\subsection{Expert Management in Edge Scenarios}

Due to sparse activation characteristics, MoE models are well suited for edge scenarios. Unlike traditional Transformer models, MoE models do not activate all parameters for each token during the inference process. Instead, they activate only a small number of the most relevant experts based on the features of the input token, significantly reducing the computational load and memory consumption. Furthermore, the sparsity in MoE models naturally supports flexible scheduling of computational resources, maximizing the efficiency of heterogeneous resource usage.

While MoE offers many benefits, expert management becomes a key issue when performing fast inference in edge scenarios due to the memory and bandwidth limitations.

First, a large number of experts introduce storage problems. Edge devices typically have limited memory, making it difficult to store the weights of all experts simultaneously. For example, the NVIDIA RTX 1080Ti GPU has only 11 GB of memory. While MoE models activate only a few experts for each token, reducing the immediate computational requirements, the total number of parameters remains extremely large. This is mainly due to the multi-expert design, where each expert contains independent parameters. For example, the percentage of expert parameters in Switch Transformers can reach up to 99\%, leading to a large increase in memory requirements, further exacerbating the memory bottleneck.

Secondly, the frequent loading of experts leads to significant I/O bottlenecks, which severely impact inference latency. The bandwidth between GPU and CPU is limited, and each time an expert is loaded, substantial I/O time is required. For latency-sensitive tasks, this has a significant impact on the model's response time and user experience.

In addition, the trade-off between expert quantization and accuracy is crucial. To store more expert weights within the limited memory, low-bit quantization techniques are often used. While quantization can effectively reduce the storage overhead, it may introduce accuracy loss, which impacts inference quality. Furthermore, quantization strategies need to be flexible for different experts to balance accuracy and performance, placing higher demands on system design.

Therefore, efficient expert management has become the cornerstone for achieving low-latency and resource-efficient MoE inference in edge scenarios.

\subsection{Challenges of MoE Inference in Edge Scenarios}

The memory and bandwidth limitations in edge scenarios pose many problems for achieving efficient inference with MoE models. To optimize inference performance, we aim to solve these problems by designing an efficient expert management. However, this faces several challenges, including prefetching, caching, and quantization of experts, as well as the difference between the prefill and decoding phases. In the following sections, we discuss these challenges in detail.

\textbf{\textit{Challenges 1: High-accuracy and low-cost expert prefetching.}}
Accurate expert prefetching is a prerequisite and the most critical challenge for achieving efficient inference with MoE models in offloading scenarios. Given the limited memory, offloading experts during MoE inference is often unavoidable. It is then necessary to prefetch the required experts in parallel to avoid high latency. However, the dynamic gating mechanism of MoE models determines that the experts activated by each token depend on input features in real-time, making expert access patterns inherently difficult to predict. And limited bandwidth means that inaccurate prefetching can trigger frequent on-demand I/O operations, significantly increasing inference latency. In addition, accurate prefetching strategies often incur computational overhead that can may negate the performance benefits of prefetching.

Existing expert prefetching methods can generally be categorized into two main types: activation path-based methods and small-network-based methods.
Activation path-based methods predict which experts are likely to be activated based on observed historical expert activation patterns. However, due to the dynamic gating nature of MoE models, expert activation patterns can vary depending on different inputs, resulting in lower prediction accuracy. Results from Lina~\cite{li2023accelerating} and MoE-Infinity~\cite{xue2024moe} have demonstrated that this method tends to perform poorly in shallow layers or when the number of experts is large. 
On the other hand, neural network-based methods use a trained predictor for expert prefetching, or even replace the original gate, achieving high accuracy. For example, SiDA~\cite{du2024sida} uses a hash network for prediction and achieves a hash hit rate of more than 90\% on Switch-base-128. Similarly, Pre-Gated MoE~\cite{hwang2024pre} directly trains a new gate to replace the original gate, effectively decoupling expert selection from expert computation. However, these methods suffer from limited portability, as a predictor trained for one MoE model cannot be easily transferred to others. Additionally, these approaches often require fine-tuning to ensure inference quality, resulting in high upfront training costs.

\textbf{\textit{Challenges 2: Expert caching with high hit rate.}}
Expert caching strategies are effective in further improving hit rates based on expert predictions. The key challenges in designing a high-hit-rate caching strategy lies in the dynamic nature of expert access patterns and the limited resources.
The dynamic routing mechanism of MoE models introduces significant uncertainty and sparsity into the expert selection for each token, making it difficult to determine which experts should be cached. At the same time, limited memory in edge scenarios prevents the storage of all expert weights. It is therefore crucial to prioritise caching of frequently accessed experts within the limited cache space, while dynamically balancing access frequency and weight size.
Existing works typically employ LRU strategies or static popular expert-based caching methods. However, these methods often fail to deliver optimal performance. In particular, static caching strategies based on popular experts are incompatible with the dynamic routing nature of MoE models. Furthermore, these approaches rarely consider the layer-wise differences in caching requirements. For example, layers with high prediction accuracy have a lower dependency on cached experts, while layers with low prediction accuracy require more cached experts to compensate for prediction errors.

\textbf{\textit{Challenges 3: Overlap of computation and I/O in the pipeline.}}
While the accuracy of expert prefetching and caching is critical, the I/O overhead during these processes cannot be overlooked, as it can contribute significantly to inference latency. Ideally, I/O data loading and computational tasks should overlap efficiently to ensure that while expert weights are being loaded, computational units can continue to perform tasks, thereby minimizing idle resource time.
However, due to the bandwidth limitations, achieving efficient overlap between computation and I/O is challenging. For example, on a PC equipped with an NVIDIA RTX 3090, the average I/O time per expert of Qwen1.5-MoE is approximately 6 ms, while the average computation time per MoE block is about 13 ms. This indicates that only two experts' I/O operations can effectively overlap with computation.
To overcome this, quantization techniques are often used to reduce the I/O overhead. However, most existing work quantizes all experts to the same bit width, without taking into account the different importance of different experts. Specially, EdgeMoE~\cite{yi2023edgemoe} adjusts the quantization bit width per expert based on its sensitivity to quantization. However, this static bit-width assignment is not well suited to the dynamic routing nature of MoE models. For example, in the prefill phase, if a low-bit expert ends up processing a large number of tokens, while a high-bit expert processes only a few tokens, the static quantization assignment may result in suboptimal inference quality or even unusable outputs.

\textbf{\textit{Challenges 4: Differences between the prefill and decoding stages of MoE inference.}}
When performing MoE inference, the differences between the prefill and decoding stages present an often overlooked challenge. These two stages exhibit differences in computational characteristics, expert activation, and resource allocation requirements, making it difficult for a unified optimization strategy to efficiently address both.
During the prefill stage, the model processes a large number of input tokens simultaneously, requiring the activation of multiple experts at once. This stage places high demands on bandwidth, memory, and computational resources. However, since prediction results from different tokens may overlap, minor prediction errors in expert selection are often mitigated by the predictions from other tokens. Therefore, the focus during the prefill stage is primarily on overlapping computation with I/O, rather than achieving absolute prediction accuracy.
In contrast, the decoding stage follows an autoregressive process, where only one token is generated at each step. Each token generated can trigger completely different expert access paths, making accurate prefetching difficult. As a result, the decoding stage prioritizes accurate expert prediction over I/O computation overlap.
Unfortunately, existing works on MoE inference rarely consider these phase-specific optimizations explicitly, often adopting uniform strategies across both stages, which fail to exploit their unique characteristics effectively.

\section{Fate Design}

\subsection{Overview}
\begin{figure}
  \centering
  \includegraphics[width=0.9\linewidth]{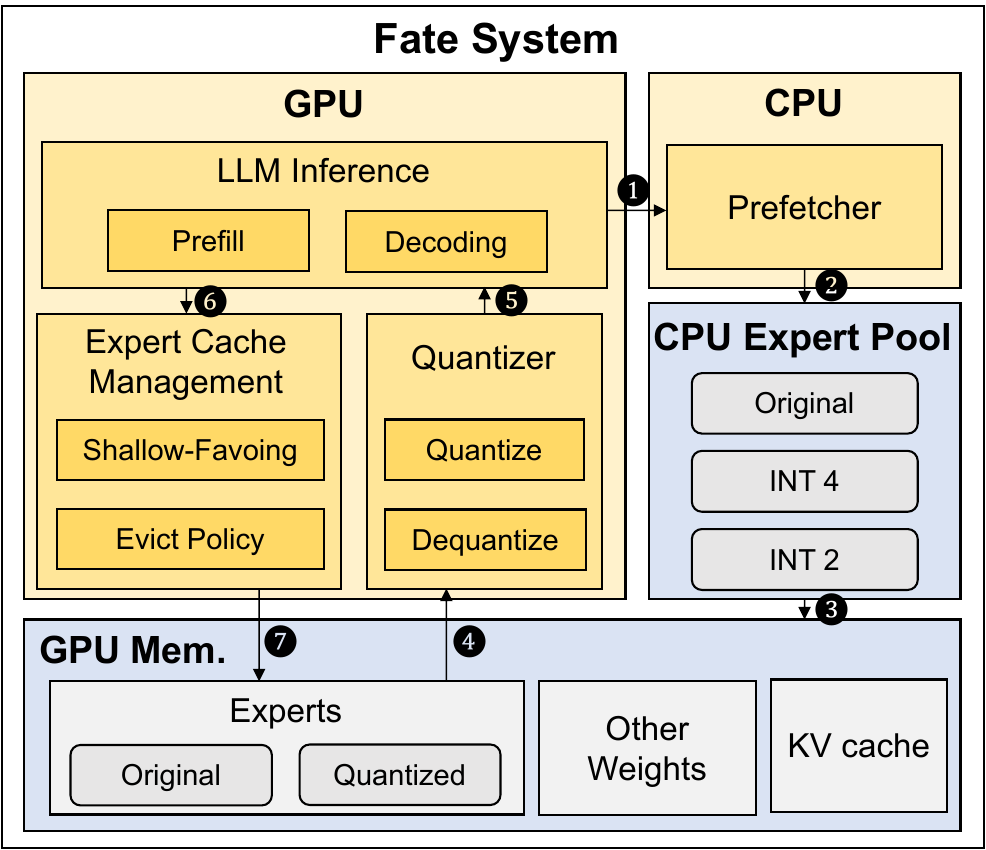}
  \caption{Overview of Fate.}
  \label{figure:overview}
\end{figure}

To tackle the above challenges, we propose Fate, which aims at fast inference of MoE models on the edge. We show the system overview of Fate in \autoref{figure:overview}.

During the inference of MoE models, Fate adopts different strategies for the prefill and decoding phases to cope with their different resource requirements (\raisebox{-0.15ex}{\ding{182}}). 
Before the gate computation in each MoE block, the input (intermediate state) is cloned to the CPU for parallel prediction computation (\raisebox{-0.15ex}{\ding{183}}). This operation generates a prefetch list containing the indexes of the experts likely to be involved in the computation at the next layer, together with the quantization bit width for each expert.
For each expert in the prefetch list, the CPU checks whether the expert is already exists in GPU memory. If an expert is not found in GPU memory, it is transferred from the CPU expert pool to GPU memory with the corresponding quantization bit width (\raisebox{-0.15ex}{\ding{184}}, \raisebox{-0.15ex}{\ding{185}}).
The GPU memory stores the dense parts of the model, the KV cache and the cached experts. When the computation reaches the next layer on the GPU, the required experts are retrieved from the GPU memory (\raisebox{-0.15ex}{\ding{186}}), and the quantized experts are dequantized before being used in the computations (\raisebox{-0.15ex}{\ding{187}}).
After completing the computations for the MoE layer, Fate updates the cache list based on the results of the gate (\raisebox{-0.15ex}{\ding{188}}). The Fate evicts experts that are not in the cache list to optimize memory usage (\raisebox{-0.15ex}{\ding{189}}).

In the following sections, we will provide a detailed explanation of each component of the aforementioned process.
First, we discuss the observed opportunities for expert prefetching (Section \ref{4.2}). Based on the findings, we propose a cross-layer expert prefetching strategy to optimize resource utilization across different layers (Section \ref{4.3}).
Next, to further improve expert cache hit rates, we introduce the shallow-favoring expert cache management strategy along with its eviction policy, which prioritizes caching strategies based on layer-specific requirements (Section \ref{4.4}).
Finally, to reduce I/O requirements while maintaining accuracy, we apply an expert-tailored quantization strategy that balances computational efficiency and model accuracy  (Section \ref{4.5}).

\subsection{Prefetching Opportunities}
\label{4.2}

\textbf{Start Point: Cosine Similarity.}
Fate's prefetching strategy is mainly based on the key observation that the inputs of neighbouring gates are very similar. Next, we provide real data to support for the effectiveness of this prefetching strategy.

Inspired by InfiniGen~\cite{lee2024infinigen}, we investigate the hidden states at three specific positions. Specifically, during inference, we focus on adjacent MoE blocks and calculate the cosine similarity between $Gate\_in_{i+1}$ and the following inputs: $Attn\_in_{i+1}$ (position 1), $Gate\_in_i$ (position 2) and $Attn_{in\_i}$ (position 3).
The results, as presented in~\autoref{table:cosine}, indicate that the average cosine similarity exceeds 83\%, with a clear trend showing that the closer the position is to $Gate\_in_{i+1}$, the higher the cosine similarity.

\begin{table}[]
\renewcommand{\arraystretch}{1.2}
\centering
\caption{Average cosine similarity and prefetch accuracy between $Gate\_in_{i+1}$ and the following inputs: $Attn\_in_{i+1}$, $Gate\_in_i$ and $Attn_{in\_i}$.}
\begin{tabular}{cccc}
\Xhline{1pt}
\multirow{2}{*}{\textbf{Position}} & \multirow{2}{*}{\textbf{Average Similarity}} & \multicolumn{2}{c}{\textbf{Prefetch Accuracy}} \\ \cline{3-4} 
 &  & \textbf{Prefill} & \textbf{Decoding} \\ \hline
1 & 93.05\% & \multirow{3}{*}{\textgreater 99.6\%} & 83.99\% \\
2 & 88.83\% &  & 78.79\% \\
3 & 83.84\% &  & 76.94\% \\
\Xhline{1pt}
\end{tabular}
\label{table:cosine}
\end{table}

Then, we further try to use $Attn\_in_{i+1}$, $Gate\_in_i$, and $Attn\_in_i$ for expert prediction. The average prediction accuracy is also shown in~\autoref{table:cosine}.
During the prefill stage, a large number of tokens are processed simultaneously, and each token selects its top-k experts. The overlap and complementarity among these prediction results lead to an accuracy close to 100\%.
In contrast, during the decoding stage, only a single token is processed at each step, making the high prediction accuracy more meaningful. The results show that during the decoding stage, the lowest prediction accuracy reaches an impressive 76.94\%. This result potentially surpasses the accuracy of some activation-aware prediction methods.

Among the three positions of the inputs, considering the trade-off between I/O time and prefetch accuracy, we choose $Gate\_in_i$ at position 2.
Although $Attn\_in_{i+1}$ achieves the highest prefetch accuracy, the computation time of a single attention layer is short, making it difficult to complete the I/O for all prefetched experts. On the other hand, selecting $Attn_{in\_i}$ offers a longer I/O time, but $Attn\_in_i$ and $Gate\_in_{i+1}$ are separated by more than one MoE block. This separation introduces a risk of I/O conflicts between adjacent layers. The I/O for experts in layer $i$ may not complete before layer $i+1$ begins its I/O. In addition, $Attn_{in\_i}$ has the lowest prediction accuracy.
Therefore, we select $Gate\_in_i$, achieving a prefetch accuracy of 78.79\% without additional training or statistical costs.

\textbf{Similar Routing Weights.}
Although the above method is simple and effective, it still results in a misprediction rate of over 20\%.
We know that I/O is one of the main causes of high latency in inference. Each misprediction requires an on-demand transfer of experts, which is unacceptable in latency-sensitive edge scenarios. Therefore, to further improve prefetch accuracy, we delve deeper into the routing weights obtained from the gate computation.

\begin{figure}[]
    \centering
    \begin{subfigure}[b]{\linewidth}
        \centering
        \includegraphics[width=\linewidth]{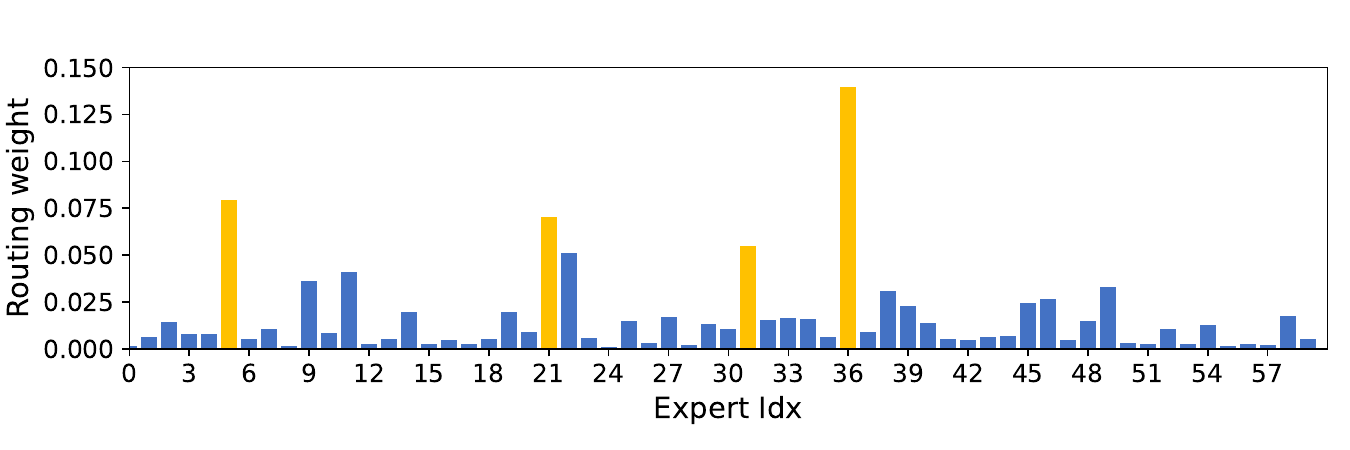} 
        \caption{The true routing weights using $Gate\_in_{i+1}$.}
        \label{fig:routinga}
    \end{subfigure}
    
    \begin{subfigure}[b]{\linewidth}
        \centering
        \includegraphics[width=\linewidth]{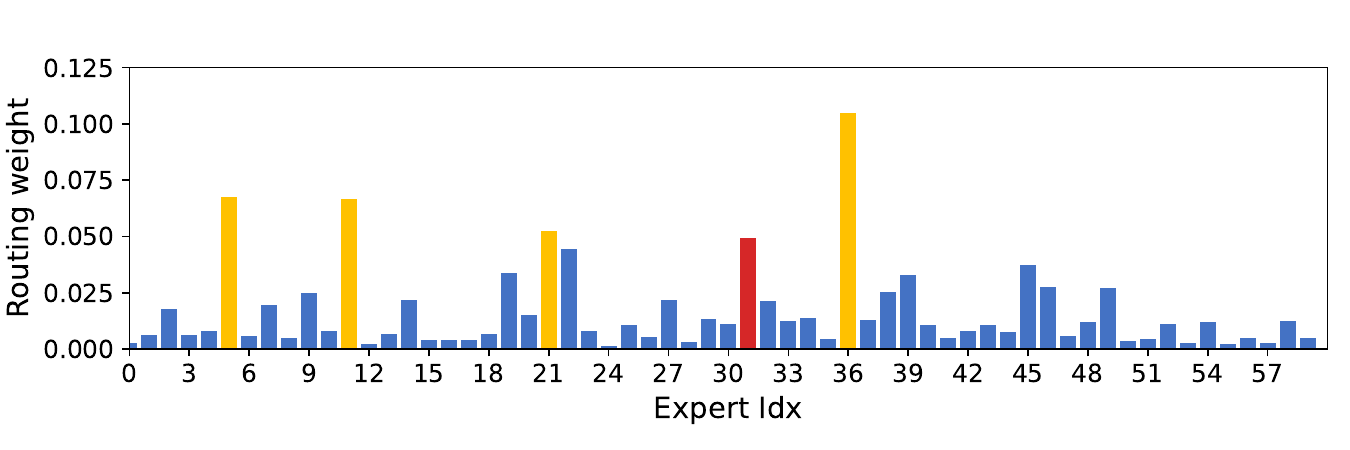} 
        \caption{The predicted routing weights using $Gate\_in_i$.}
        \label{fig:routingb}
    \end{subfigure}

    \caption{Comparison between true and predicted routing weights at block $i+1$.}
    \label{figure:routing}
\end{figure}

The routing weights computed by the gate include the probability that each expert is selected. The sum of the probabilities across all experts is 1. The top-k experts are selected based on the highest routing weights as the final output.
We randomly select results of once misprefetch, and compare the distribution of routing weights obtained from the predictions with those obtained from the actual computations. The true results are shown in~\autoref{fig:routinga} and predicted results are shown in~\autoref{fig:routingb}. It is evident that the routing weight distributions in both figures are remarkably similar. However, the actual selected experts are [5, 21, 31, 36], whereas the predicted results are [5, 11, 21, 36], with the missing expert 31 ranked fifth in the prediction results.
This observation suggests that under similar routing weight distributions, increasing the number of prefetched experts could likely include the truly required experts.
Based on this insight, we attempted to transfer experts whose confidence scores exceed the 75th percentile, achieving an impressive prefetch accuracy of 97.15\%.

\subsection{Cross-Layer Expert Prefetch}
\label{4.3}

\begin{figure}
  \centering
  \includegraphics[width=\linewidth]{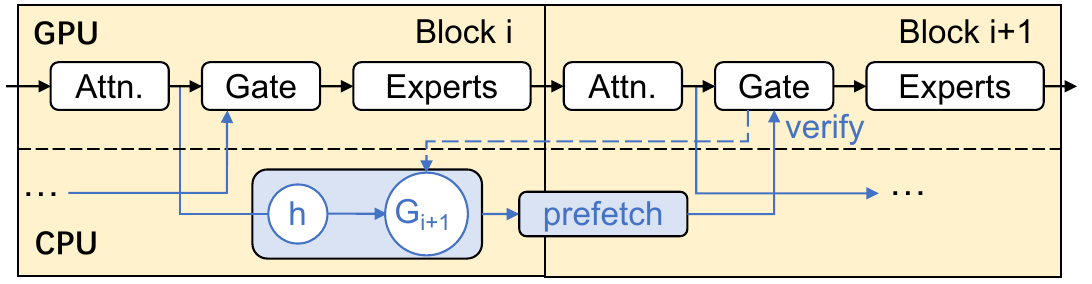}
  \caption{Workflow of cross-layer expert prefetch.}
  \label{figure:prefetch}
\end{figure}

Building on the prefetching opportunities observed in Section \ref{4.2}, the workflow of expert prefetching in Fate is illustrated in~\autoref{figure:prefetch}.
At the beginning of gate in the $i$-th block, Fate clone the input to the CPU and feed it into the gate of the ($i+1$)-th block in parallel. The results are then used to prefetch the experts.
This entire prediction process runs in parallel with GPU computations, achieving exceptionally high accuracy without introducing any additional overhead to the GPU inference process. Furthermore, because the sizes of the gate and the input are small, the computational burden on the CPU remains minimal.
This strategy is remarkably elegant due to its simplicity, effectiveness, and broad applicability across different scenarios.

\begin{figure}
  \centering
  \includegraphics[width=\linewidth]{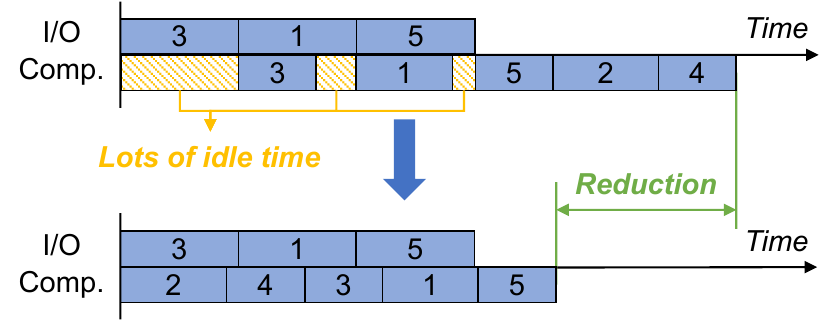}
  \caption{Example of rearranging the order of expert computations in the prefill stage. Have prefetched: 2, 4. Loading: 3, 1, 5. Original order: 3, 1, 5, 2, 4. Our order: 2, 4, 3, 1, 5.}
  \label{figure:prefill}
\end{figure}

\textbf{Prefill Stage.} During the prefill stage, where all tokens in the sequence are processed simultaneously, a large number of experts are activated. Therefore, Fate focuses on minimizing pipeline bubbles between expert computations. As illustrated in~\autoref{figure:prefill}, a naive approach sequentially transfers experts and computes based on the order of token requests, which can lead to significant idle time in the inference pipeline.
To address this, Fate statistically aggregates token-to-expert assignments after obtaining expert prediction results, calculating the popularity of each expert based on their frequency of selection across all tokens. Experts are then transferred in descending order of popularity, prioritizing those involved in computations for a larger number of tokens. This approach ensures that the most frequently used experts are available earlier, providing additional time for subsequent expert I/O, thus improving pipeline efficiency.

\textbf{Decoding Stage.} During the decoding stage, where only a single token is processed at a time, the activated experts are limited to the top-k. At this stage, Fate prioritizes achieving full overlap between computation and I/O.  
First, Fate statistically pre-computes the average computation times for each layer during decoding, including \( t_{moe} \), \( t_{attn.} \), and \( t_{gate} \), which represent the computation time of the MoE layer, gate, and attention, respectively. Using these values, the time constraint \( T \) can be derived as:
\[
T = t_{moe} + t_{attn.} + t_{gate}
\]
In addition, Fate also measures the I/O time for a single expert (\( t_{expert} \)). Based on the time constraint $T$, the maximum number of experts that can be transferred is calculated as follows:
\[
n = \frac{T}{t_{expert}}
\]
Importantly, this calculation of \( n \) is performed offline, ensuring that no additional computational overhead is introduced during the inference process. By pre-computing this, Fate ensures seamless overlap between computation and I/O during decoding, effectively minimizing latency.

\subsection{Shallow-Favoring Expert Cache}
\label{4.4}

To further improve expert hit rates, Fate leverages available memory resources to cache experts, thereby reducing the I/O overhead. However, in real-world deployment scenarios, memory resources are often shared among multiple tasks, not just dedicated to LLM inference. And the memory size of different devices can vary.
To address this, Fate defines a memory budget to limit the peak memory usage during inference. Fate then models the memory usage based on the memory footprint of each layer, KV cache and activation, etc. After subtracting the memory occupied by dense parts, the remaining memory is allocated for expert caching. 
The total number of experts that can be cached (\( cache\_total \)) is derived using the following formula:
\[
cache\_total = \frac{memory\_budget - others}{expert}
\]
where \( memory\_budget \) refers to the total memory allocated for inference. \( others \) refers to memory consumed by non-expert parts (e.g., dense layers, KV cache, etc). \( expert \) refers to the memory required per expert.  
By enforcing a memory budget and dynamically allocating cache space, Fate ensures optimal memory utilization while minimizing I/O latency during expert retrieval.

\begin{figure}
  \centering
  \includegraphics[width=\linewidth]{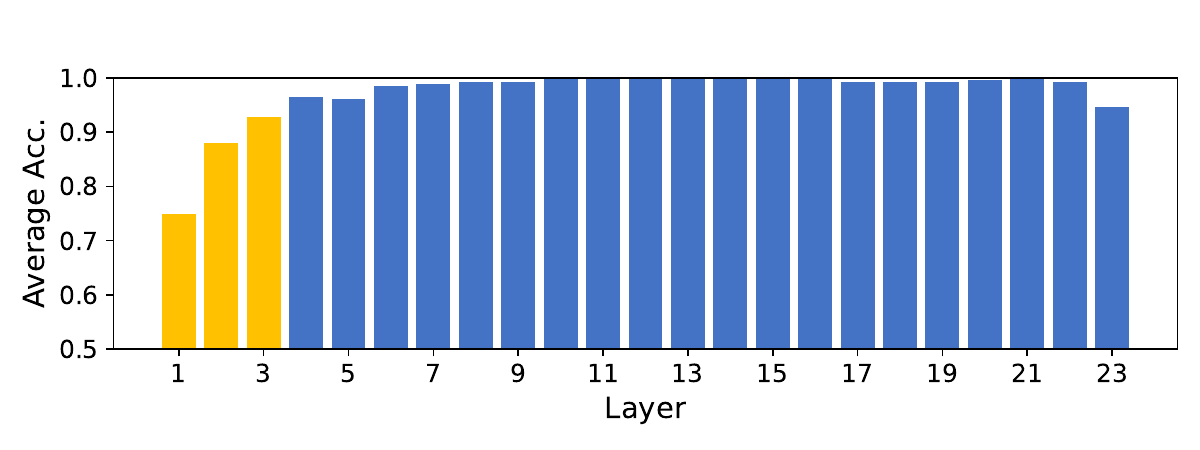}
  \caption{Average accuracy of expert prefetching at each layer.}
  \label{figure:avg_acc}
\end{figure}

\begin{figure}[]
    \centering
    \begin{subfigure}[b]{\linewidth}
        \centering
        \includegraphics[width=\linewidth]{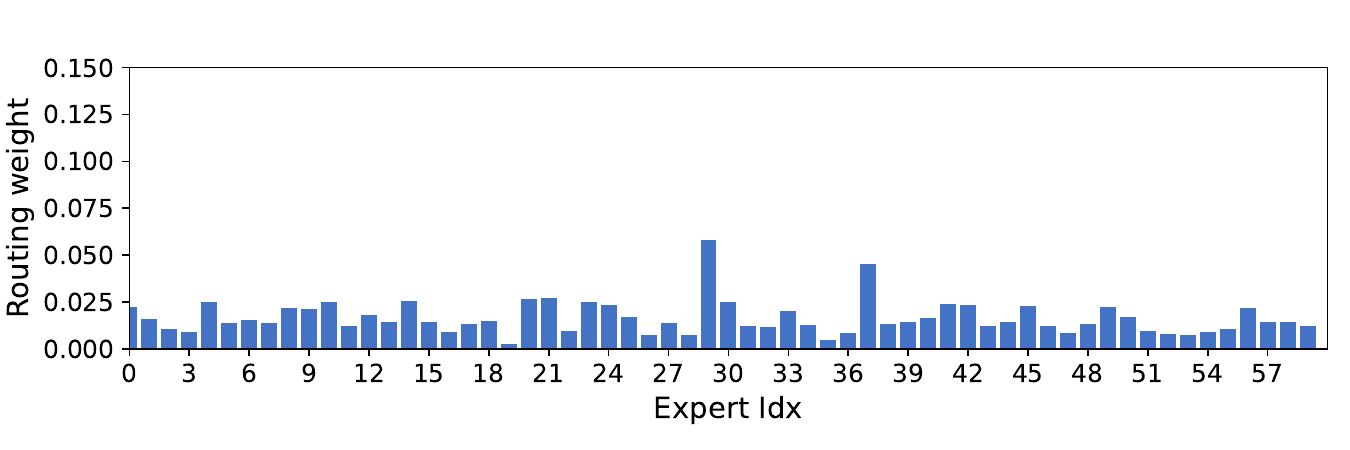}
        \caption{Routing weights of the first layer.}
        \label{fig:first}
    \end{subfigure}
    
    \begin{subfigure}[b]{\linewidth}
        \centering
        \includegraphics[width=\linewidth]{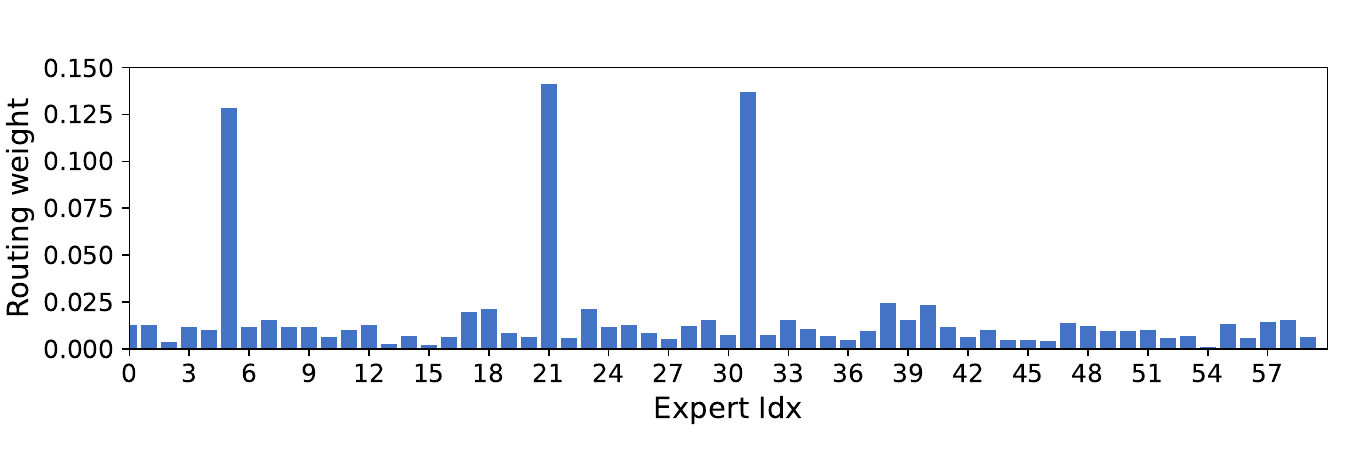}
        \caption{Routing weights of the 23rd layer.}
        \label{fig:23rd}
    \end{subfigure}

    \caption{Routing weights of the first layer and the 23rd layer.}
    \label{figure:routing_wei}
\end{figure}

In order to allocate $cache\_total$ more effectively across different layers, we have calculated the average prefetch accuracy for each layer based on the findings in Section~\ref{4.2}. As shown in~\autoref{figure:avg_acc}, the prefetch accuracy in shallow layers is obviously lower than in deeper layers.
To understand this discrepancy, we further examined the differences in routing weights between shallow and deep layers, as shown in~\autoref{figure:routing_wei}. Comparing~\autoref{fig:first} and~\autoref{fig:23rd}, it is clear that in shallow layers, token routing preferences are less distinct across experts, with probabilities more evenly distributed across experts. In contrast, deeper layers show a clear routing preference, where tokens are more decisively assigned to specific experts. As a result, predictions using inputs with high cosine similarity achieve higher accuracy in deeper layers.
Motivated by this observation, we conducted a simple experiment: we cached all experts in layers 0-3. The results showed an average expert hit rate of 99.08\%, demonstrating the effectiveness of targeted caching in shallow layers.

Based on the experimental results, Fate introduces the shallow-favoring expert cache strategy. Specifically, Fate sets a boundary \( L \) between shallow and deep layers based on offline statistical analysis, and prioritises the allocation of memory resources to shallow layer experts, while the remaining memory is distributed evenly across the deep layers. For example, from the observed data in~\autoref{figure:avg_acc}, the boundary for Qwen1.5-MoE is determined as \( L = 3 \). If memory resources are limited, Fate allocates expert cache space sequentially to the first \( L \) layers. If memory resources are sufficient, all shallow layer experts are fully cached, and the cache space for each deep layer is calculated as:  
\[
cache\_deep = \frac{cache\_total - L \times num\_experts}{num\_layers - L}
\]
where \( num\_experts \) is the number of experts per layer, and \( num\_layers \) is the total number of layers in the model.  

To manage expert caching within each layer, Fate uses the Adaptive Replacement Cache (ARC) algorithm for expert eviction. While the commonly adopted approach is LRU caching, observations of the popular experts~\cite{li2023accelerating} indicate that activation frequency is also a critical factor in determining the importance of an expert during MoE inference. ARC dynamically balances both recency and frequency, ensuring efficient cache management under varying access patterns.
The ARC algorithm maintains four key queues: T1 (recently accessed cache experts), T2 (frequently accessed cache experts), B1 (experts evicted from T1), and B2 (experts evicted from T2). By dynamically adjusting the sizes of T1 and T2, ARC can adaptively switch strategies in response to changing access patterns. For example, in short-term hotspot scenarios, ARC expands T1 to prioritize recently accessed experts, ensuring immediate availability. Conversely, in long-term or frequent access scenarios, ARC prioritizes T2, preserving high-frequency experts for sustained reuse. This adaptive balancing mechanism allows ARC to outperform other static strategies, providing robust performance in the face of complex and evolving expert access patterns in inference.

\subsection{Expert Quantization Management}
\label{4.5}

To accelerate I/O operations and reduce the memory footprint of expert caching, Fate applies quantization to experts that consume significant memory resources. After evaluating a range of popular quantization algorithms, we selected HQQ (Half-Quadratic Quantization~\cite{badri2023hqq}), which has been demonstrated to achieve excellent performance on MoE models while offering fast quantization and dequantization speeds.

\begin{table}[]
\centering
\caption{Impact of quantization on the inference accuracy of Qwen1.5-MoE. "Acc.": accuracy; "Degrad.": accuracy degradation compared to original accuracy. Benchmark: MMLU.}
\begin{tabular}{ccccc}
\Xhline{1pt}
\multirow{2}{*}{\textbf{Model}} & \multicolumn{4}{c}{\textbf{Qwen-1.5-MoE}} \\ \cline{2-5} 
 & \textbf{Original} & \textbf{INT8} & \textbf{INT4} & \textbf{INT2} \\ \hline
Acc. & 59.6 & 59.6 & 59.4 & 55.1 \\
Degrad. & 0 & 0 & -0.2 & -4.5 \\
\Xhline{1pt}
\end{tabular}
\label{quan}
\end{table}

\textbf{Quantization for Cache.} During expert caching, Fate employs int4 quantization when memory resources are insufficient. Existing research indicates that MoE experts exhibit strong robustness to quantization~\cite{kim2023mixture}. As shown in~\autoref{quan}, when we applied INT8 quantization to all experts in the Qwen1.5-MoE model, we observed minimal accuracy loss. Even with INT4 quantization, the accuracy degradation remained remarkably low.
Considering the trade-off between memory consumption and accuracy loss, we opted for INT4 quantization. This choice significantly reduces the memory footprint per expert, allowing a substantial increase in the total number of cached experts ($cache\_total$).
When prefetched experts are cached in INT4 format, Fate performs dequantization, which is faster than directly loading the experts.

\begin{figure}
  \centering
  \includegraphics[width=\linewidth]{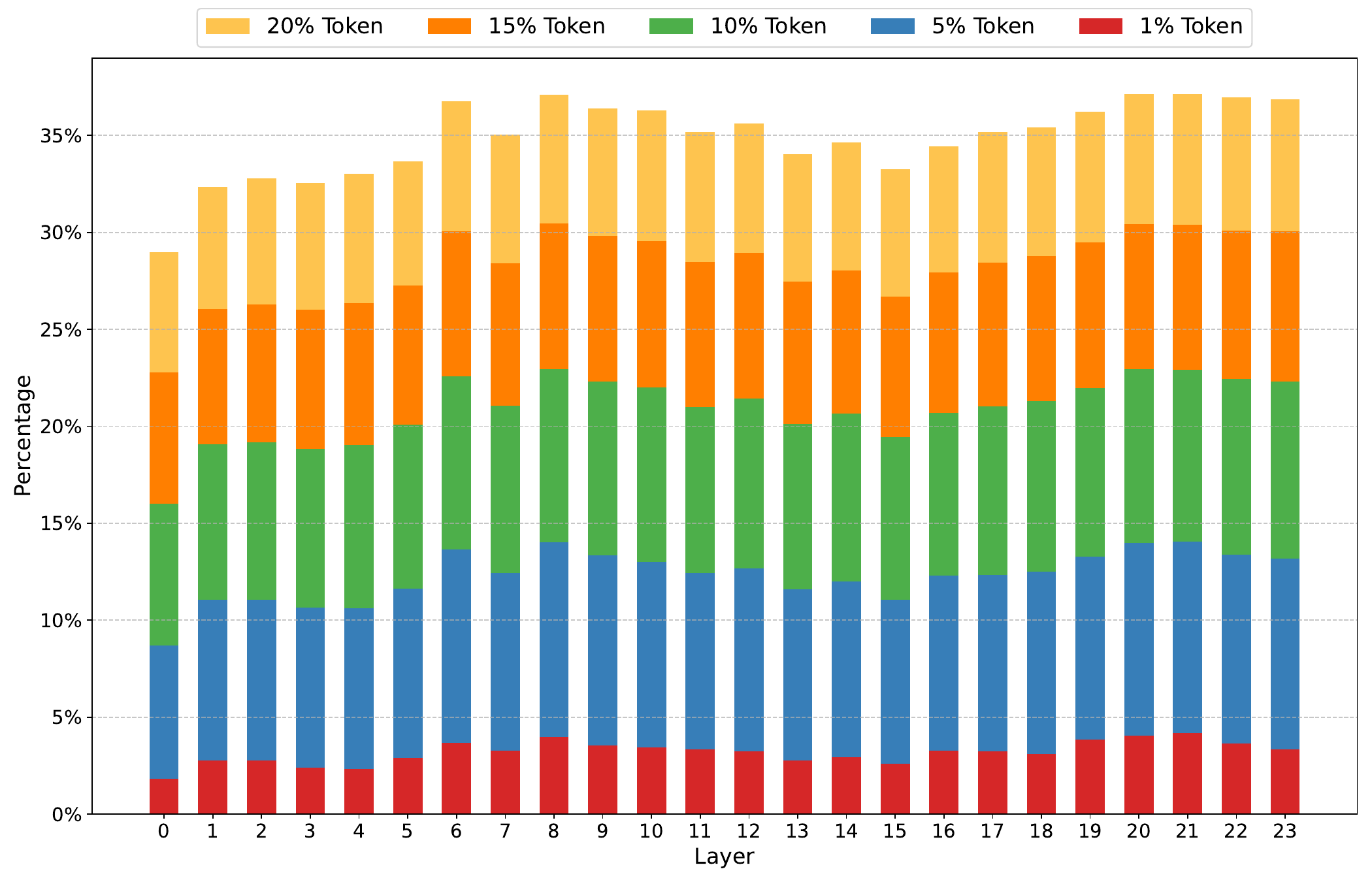}
  \caption{The distribution ratio of the tokens among the experts in each layer. The y-axis represents the percentage of experts that process this part of the tokens.}
  \label{figure:token_distribution}
\end{figure}

\textbf{Quantization for I/O.} During the expert transfer, Fate employs different quantization strategies tailored to the characteristics of the prefill and decoding stages.
In the prefill stage, Fate introduces a popularity-aware hybrid quantization strategy. In this stage, all tokens in the input sequence are processed simultaneously, resulting in the activation of a large number of experts. For example, using the MMLU dataset, we observed that in the prefill stage of Qwen1.5-MoE, an average of 55.9 experts are activated per layer out of a total of 60 experts. This stage faces significant I/O pressure.
However, treating all experts equally during I/O may not be optimal, as the popularity of each expert (measured by the number of tokens processed by each expert) varies significantly. As shown in~\autoref{figure:token_distribution}, we ranked all experts in descending order of popularity and analyzed the distribution of tokens among the across non-popular experts.
The results show a clear imbalance between the number of non-popular experts and the tokens they cover. For instance, in layer 6, the orange section indicates that 30\% of the active experts cover only 15\% of the tokens. Therefore, it may not be worth transferring these non-popular experts for the same I/O cost.

\begin{figure}
  \centering
  \includegraphics[width=\linewidth]{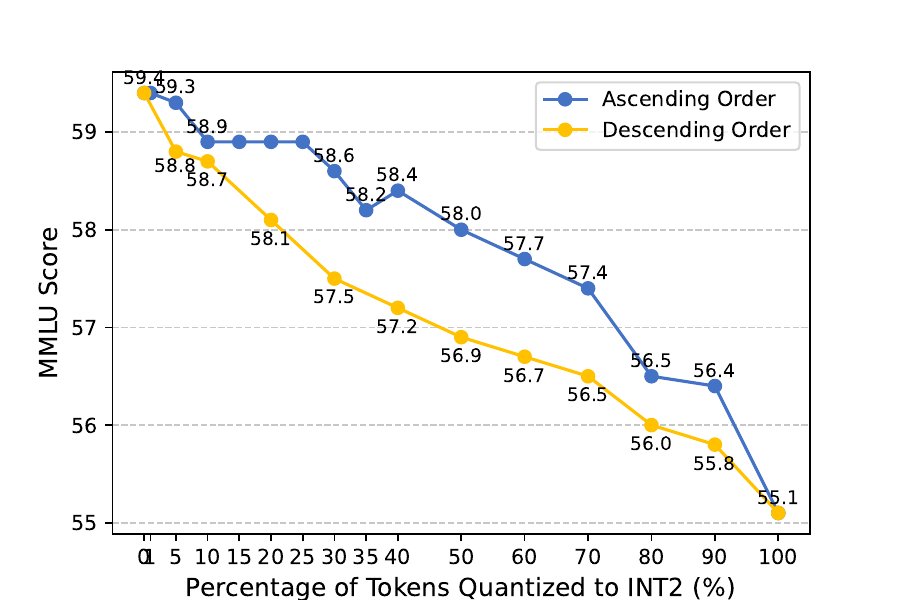}
  \caption{The trend of MMLU scores as the percentage of INT2 experts increases. The blue line shows the experts quantized in reverse order of popularity, and the yellow line shows the situation in positive order. Model: Qwen1.5-MoE.}
  \label{figure:Accuracy}
\end{figure}

Furthermore, we quantize experts covering different proportions of tokens into INT2, as illustrated in~\autoref{figure:Accuracy}. It is evident that when experts are quantized in descending order of popularity, converting non-popular experts from INT4 to INT2 has minimal impact on inference quality. In contrast, quantizing popular experts to INT2 results in a significant drop in performance scores.
Based on this, Fate stores both INT2 and INT4 versions of all experts in CPU memory, which does not place a significant memory load on the CPU memory. During the prefill stage, Fate dynamically transfers experts in either INT2 or INT4 format based on their predicted popularity.
For different models, Fate uses a heuristic search strategy to determine the optimal proportion $p$ of INT2 experts, ensuring that the loss in accuracy is within 1\%. For instance, as shown in the figure, for Qwen1.5-MoE, the optimal value is determined to be $p=25\%$.
On the other hand, during the decoding stage, since only one token is processed at a time, each selected expert is equally important. Therefore, Fate only transfers the INT4 version of each expert.

\section{Evaluation}

\subsection{Experimental Setup}

\textbf{Hardware.} 
We ran experiments on two different PC configurations to evaluate Fate's effectiveness in different environments, representing the high-end and low-end hardware scenarios of edge PCs. 
The high-end PC is equipped with an NVIDIA RTX 3090 GPU (24G), an Intel i9-9900X CPU and 128GB of host memory. PCIe 3.0 ×16 (16GB/s bandwidth) interconnects the CPU and GPU.
On the other hand, the low-end PC is equipped with an NVIDIA RTX 1080Ti GPU (11G), an Intel E5-2650 v4 CPU and 64GB of host memory. PCIe 1.0 ×16 (4GB/s bandwidth) interconnects the CPU and GPU.

\noindent
\textbf{Models.} 
We use DeepseekMoE (16.4B) and Qwen1.5-MoE (14.3B), which are popular MoE models. Both use the latest shared expert and fine-grained expert structures and perform well on various benchmarks. DeepseekMoE's gate selects the top 6 out of 64 experts and Qwen1.5-MoE's gate selects the top 4 out of 60 experts.

\noindent
\textbf{Workload.}
We use representative datasets as workloads to evaluate both performance and accuracy.
For end-to-end performance evaluation, we use ChatGPT-prompts, HumanEval, and GSM8K, which represent typical LLM applications such as chatbots, code generation, and mathematical problem solving, respectively. We measured prefill performance with input lengths of 128, 256, and 512, while decoding performance is evaluated with an input length not exceeding 64 and an output length not exceeding 1024.
For inference accuracy evaluation, we use three widely recognized benchmarks: MMLU, GSM8K, and HumanEval, to analyze the impact of Fate's quantization strategy on accuracy.

\noindent
\textbf{Baseline.} 
We compare Fate with two baselines.
(1) Load on Demand (LoD): This approach keeps only the dense components of the model in GPU memory, including shared experts. Experts are dynamically loaded based on routing results from the gate layer.
(2) Expert activation path-based strategy (EAP): We implement this method using a caching strategy identical to Fate, but using a common activation-aware expert prefetching approach. The expert choice for the next block is predicted based on the token's expert choices in the previous block and the corresponding statistical data.

\noindent
\textbf{Key Metrics.}
Our primary focus is on achieving fast inference of MoE models. Therefore, we evaluate end-to-end performance using two key metrics: prefill speed (tokens/s) and decoding speed (tokens/s). In addition, we use inference accuracy to evaluate the impact of quantization.

\subsection{End-to-End Performance}

\subsubsection{Prefill Performance}

\begin{figure}
  \centering
  \includegraphics[width=\linewidth]{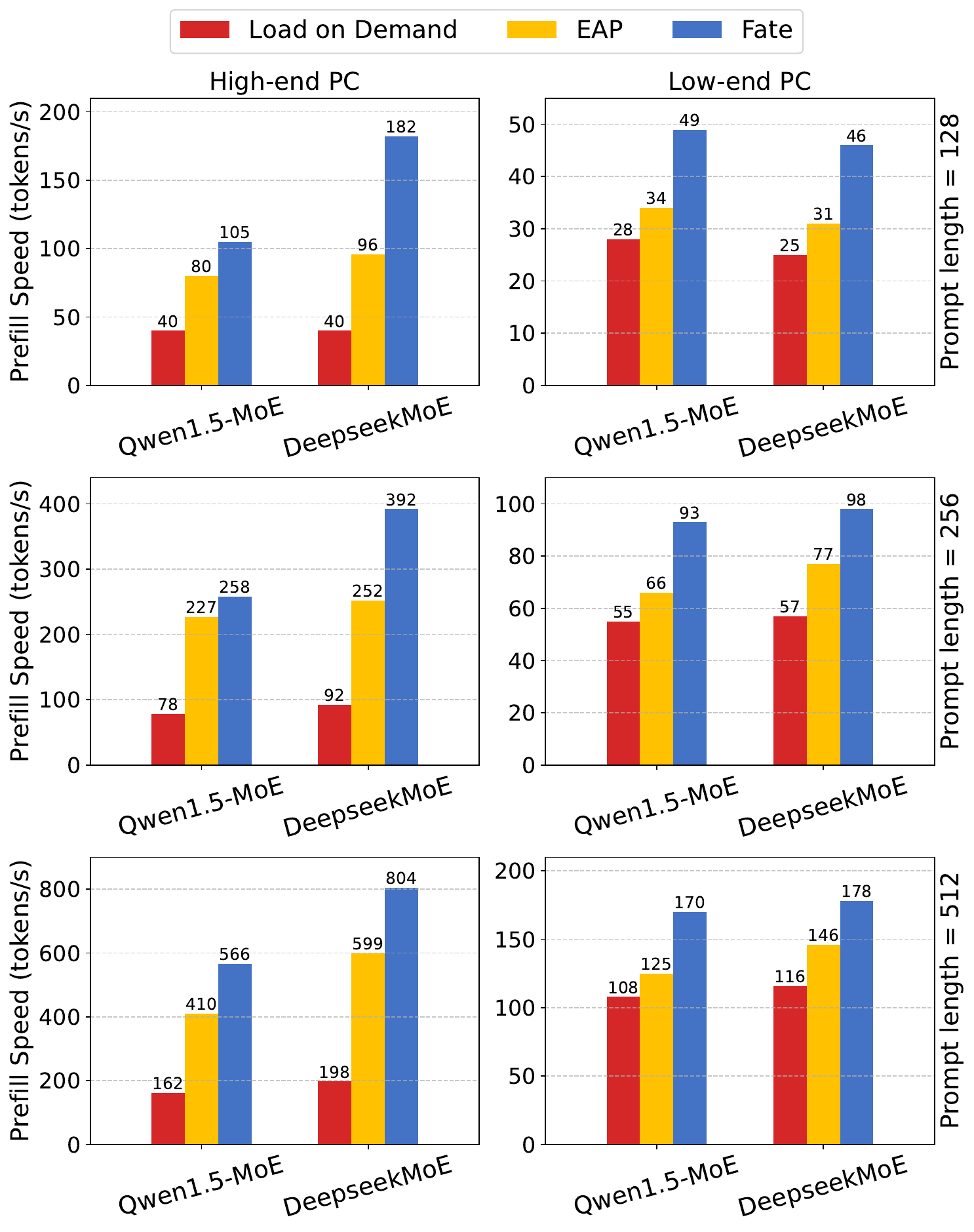}
  \caption{Prefill speeds of Fate, EAP and Load on Demand in various offloading scenarios at prompt lengths of 128, 256, and 512 tokens.}
  \label{figure:prefill_speed}
\end{figure}

We first evaluate Fate's prefill performance across different prompt lengths (128, 256, 512) compared to baselines, as shown in~\autoref{figure:prefill_speed}. Across different PCs, models, and prompt lengths, Fate shows significant performance advantages over other strategies. On the high-end PC, Fate achieves up to a 4.5× speedup over Load on Demand and a 1.9× speedup over EAP. Specifically, Fate achieves a throughput of up to 804 tokens/s with a prompt length of 512 for DeepseekMoE. On a low-end PC, Fate achieves up to a 1.8× speedup over Load on Demand and a 1.5× speedup over EAP. The lower speedup on the low-end PC is primarily due to the outdated PCIe version, which results in longer I/O times and limits overlap efficiency. However, the results clearly demonstrate that Fate consistently outperforms baselines across diverse scenarios, highlighting its advanced capabilities and scalability.

The acceleration achieved by Fate is primarily due to its strategies tailored to the prefill stage. Unlike other methods that perform sequential computation at the expert layer, Fate identifies that the prefill stage of MoE inference at the edge is I/O intensive. Therefore, Fate reorders the computation sequence based on the popularity of each expert, leveraging the longer computation times of popular experts to provide extended I/O time for other experts, thereby significantly reducing the idle time between experts. This reordering process is supported by highly accurate expert prefetching, which ensures a reliable popularity-based sequence.
Moreover, to alleviate I/O pressure during the prefill stage, Fate introduces a popularity-aware hybrid quantization strategy that quantizes non-popular experts to INT2. This approach effectively reduces I/O overhead while maintaining inference quality.

\subsubsection{Decoding Performance}

\begin{figure}
  \centering
  \includegraphics[width=\linewidth]{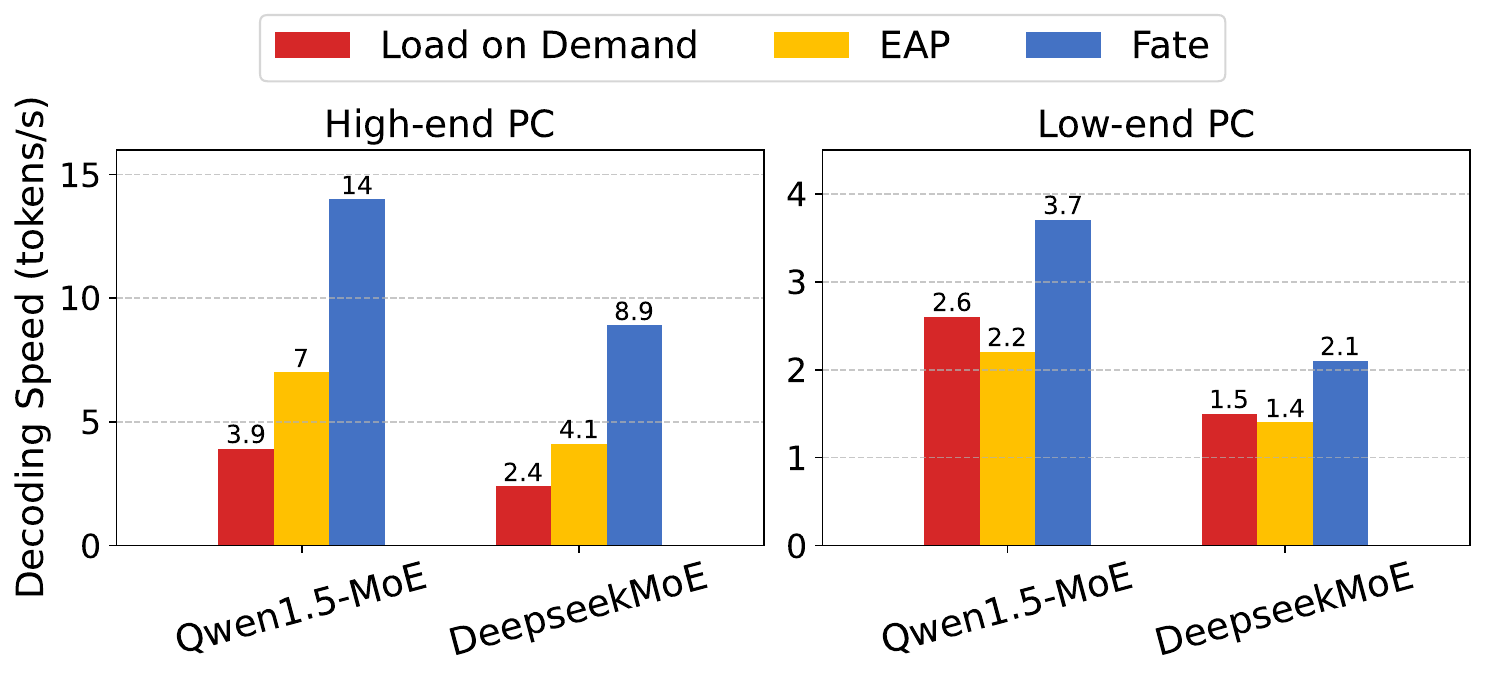}
  \caption{Decoding speeds of Fate, EAP, and Load on Demand in various scenarios.}
  \label{figure:decoding_speed}
\end{figure}

\autoref{figure:decoding_speed} illustrates the decoding speed of three strategies evaluated on two MoE models under high-end and low-end PC setups. Across all settings, Fate demonstrates a clear performance advantage over the other strategies.

On high-end PCs, Fate achieves a decoding speed of 14 tokens/s for Qwen1.5-MoE, significantly surpassing EAP and Load on Demand. For DeepseekMoE, Fate reaches a 3.7× improvement over Load on Demand and a 2.2× increase compared to EAP. These results highlight the efficiency of Fate in reducing decoding latency on high-performance hardware.
On the other hand, even though outdated PCIe severely slows down inference on low-end PCs, Fate is still significantly better than the baselines. Fate can reach up to 2.3× speedup over Load on Demand and 1.6× speedup over EAP.
Especially, on low-end PCs, the decoding speed of Load on Demand exceeds that of EAP. This is because EAP's low prediction accuracy of EAP introduces two sources of I/O latency: waiting for prefetching to complete and loading the correct experts. This results in higher I/O overhead and longer delays compared to Load on Demand. This further validates the high accuracy of Fate's prefetching strategy.

During the decoding stage, each forward processes only a single token, making decoding performance highly dependent on the effectiveness of expert prefetching and expert caching. Fate significantly improves expert hit rates by leveraging high-accuracy cross-layer expert prefetching and shallow-favoring expert caching strategies. These mechanisms effectively overlap substantial I/O overhead, reducing latency. In contrast, EAP and Load on Demand are prone to temporary I/O bottlenecks, resulting in higher inference delays and significantly lower decoding speeds than Fate. Consequently, Fate enables fast inference of MoE models in the offloading scenario.

\subsubsection{Memory Budget}

\begin{figure}[]
    \centering
    \begin{subfigure}[b]{\linewidth} 
        \centering
        \includegraphics[width=\linewidth]{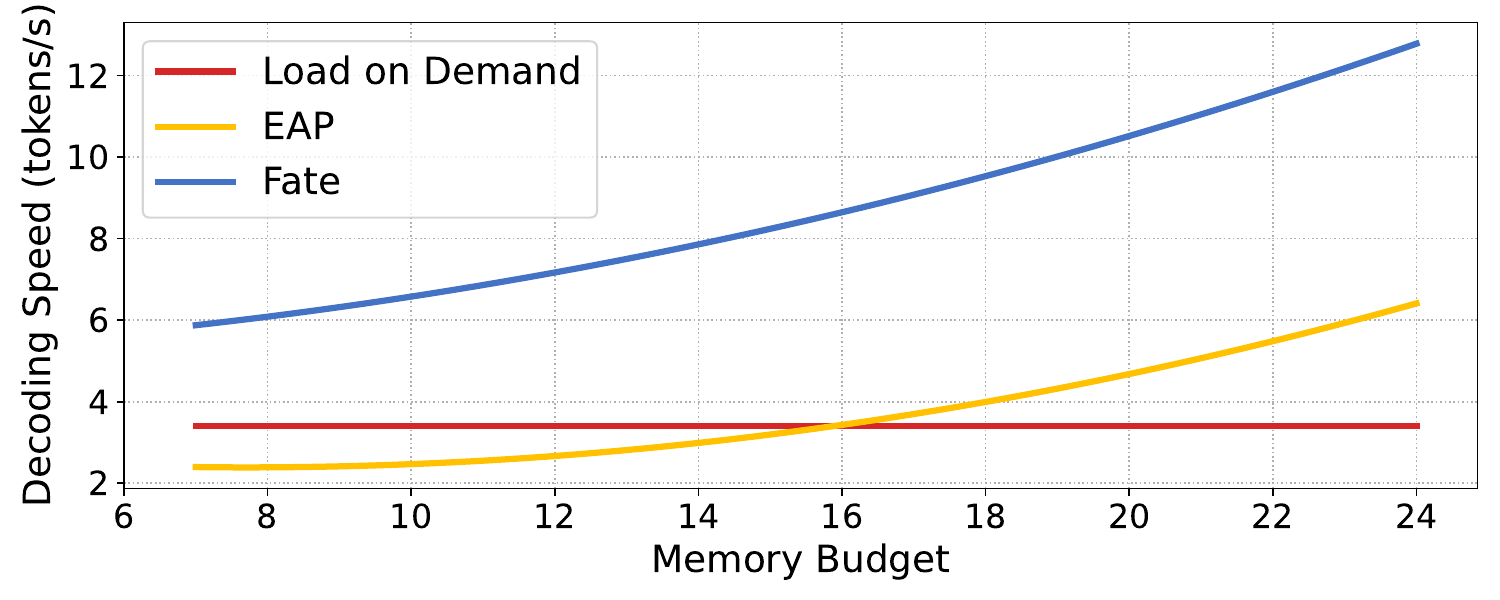}
        \caption{Qwen1.5-MoE}
        \label{fig:mem_budget_qwen}
    \end{subfigure}
    
    \begin{subfigure}[b]{\linewidth}
        \centering
        \includegraphics[width=\linewidth]{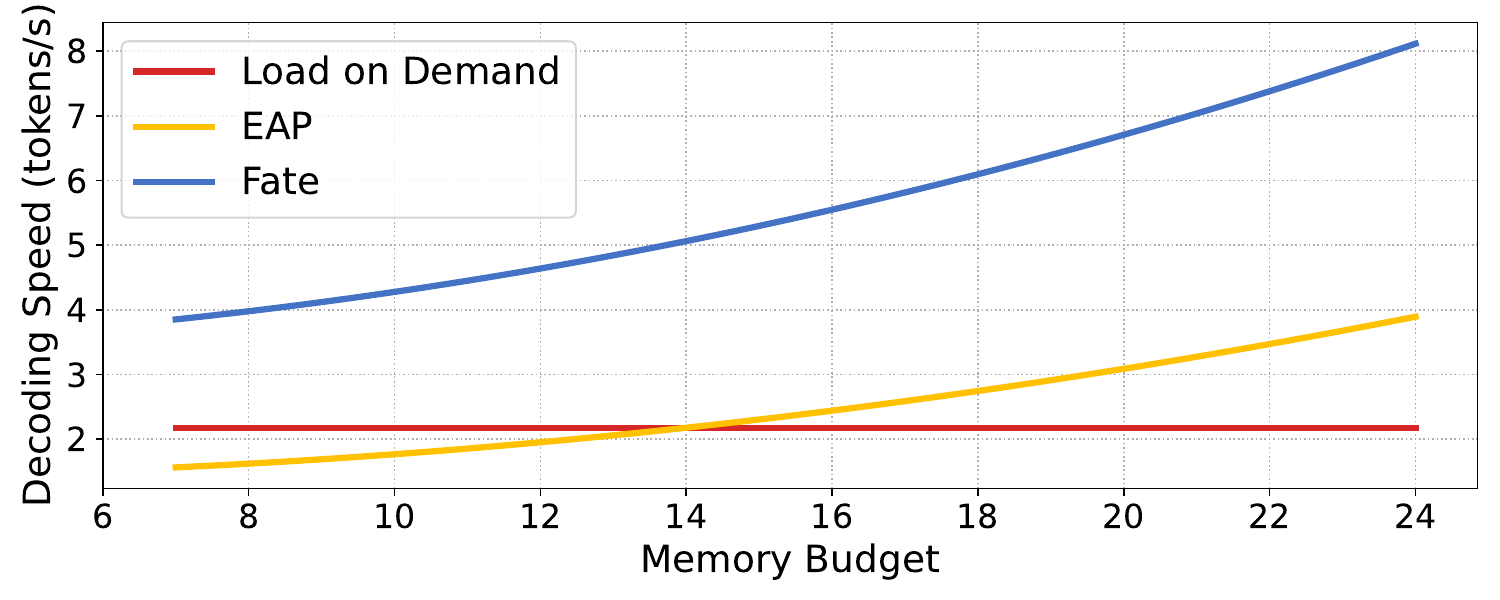}
        \caption{DeepseekMoE}
        \label{fig:mem_budget_deepseek}
    \end{subfigure}

    \caption{Decoding speed of different models in high-end PC varies with memory budget.}
    \label{figure:mem_budget}
\end{figure}

In real-world scenarios, hardware devices have varying memory capacities, and available memory is often shared with other programs rather than being dedicated exclusively to LLM inference. To evaluate the scalability and adaptability of Fate under different memory constraints, we assessed its decoding speed over a range of memory budgets (7GB to 24GB) and compared it to the baselines, as shown in~\autoref{figure:mem_budget}. Notably, the Load on Demand strategy only requires memory to store the dense part of the models, so the decoding speed remains unaffected by changes in the memory budget.

Firstly, under low memory conditions, EAP often fails due to insufficient accuracy. Even after prefetching experts, EAP may still need to load the correct experts on demand, resulting in delays that can make its performance worse than the simplest Load on Demand approach. In contrast, Fate, with its more accurate prefetching strategy, achieves a significant speedup over Load on Demand, even in scenarios where most experts are offloaded, delivering up to a 1.8× performance improvement. As memory increases, since EAP applies the same caching strategy as Fate, both maximize expert caching under the available memory, leveraging shallow-favoring expert caching strategy to further improve hit rates and reduce I/O overhead. Finally, Fate achieves the fastest MoE inference speed in offloading scenarios, delivering inference speeds of up to 14 tokens/s for Qwen1.5-MoE.

As a result, Fate demonstrates exceptional scalability and adaptability by using highly accurate prediction strategies, optimizing expert caching, and effectively overlapping computation and I/O. These capabilities enable it to deliver high-performance MoE inference across devices with different memory capacities.

\subsection{Inference Accuracy}

\begin{table}[]
\centering
\caption{The impact of Fate on the inference accuracy of two models on two PCs.}
\begin{tabular}{cccc}
\Xhline{1pt}
\textbf{MMLU} & \textbf{BF16} & \textbf{High-end PC} & \textbf{Low-end PC} \\ \hline
Qwen1.5-MoE & 59.6 & 59.7 & 58.4 \\
DeepseekMoE & 44.4 & 44.3 & 42 \\ \hline
\textbf{Avg. Degrad} & / & 0 & -1.8 \\ \hhline{====}
\textbf{GSM8K} & \textbf{BF16} & \textbf{High-end PC} & \textbf{Low-end PC} \\ \hline
Qwen1.5-MoE & 62.4 & 60.2 & 59.9 \\
DeepseekMoE & 21.2 & 22.3 & 20.2 \\ \hline
\textbf{Avg. Degrad} & / & -0.55 & -1.75 \\ \hhline{====}
\textbf{HumanEval} & \textbf{BF16} & \textbf{High-end PC} & \textbf{Low-end PC} \\ \hline
Qwen1.5-MoE & 34.8 & 32.3 & 34.1 \\
DeepseekMoE & 25.6 & 26.8 & 24.6 \\ \hline
\textbf{Avg. Degrad} & / & -0.15 & -0.85 \\
\Xhline{1pt}
\end{tabular}
\label{table:acc}
\end{table}

We evaluate the impact of Fate's quantization strategy on inference accuracy using three representative benchmarks: MMLU, GSM8K, and HumanEval. Since Fate adaptively determines the number of cached experts and their quantization bit widths based on the hardware environment, we conduct evaluations on both a high-end PC and a low-end PC. The results obtained on each benchmark are compared with the original model's accuracy scores, and the average degradation on each device is calculated. As shown in~\autoref{table:acc}, Fate achieves acceptable reductions in inference quality while maintaining high efficiency.

Specifically, Fate achieves an average accuracy loss of less than 1\% on the high-end PC, with slight improvements observed on certain benchmarks. This is due to Fate's ability to accurately identify experts that handle fewer tokens and quantize them to INT2, which not only improves I/O efficiency but also has minimal impact on inference quality. On the low-end PC, although the average accuracy loss is over 3\%, the actual score reduction is less than 2 points. The reason for this loss is that while Fate limits the number of experts quantized to INT2 during prefetch, it uses the INT2 expert for fast temporary I/O in cases of misprefetch. However, the practical reduction in score is small, and this trade-off allows an improvement in inference speed of over 2×.
Overall, these results demonstrate that Fate can effectively balance inference performance improvements with accuracy preservation.

\subsection{Ablation Study}

\begin{figure}
  \centering
  \includegraphics[width=\linewidth]{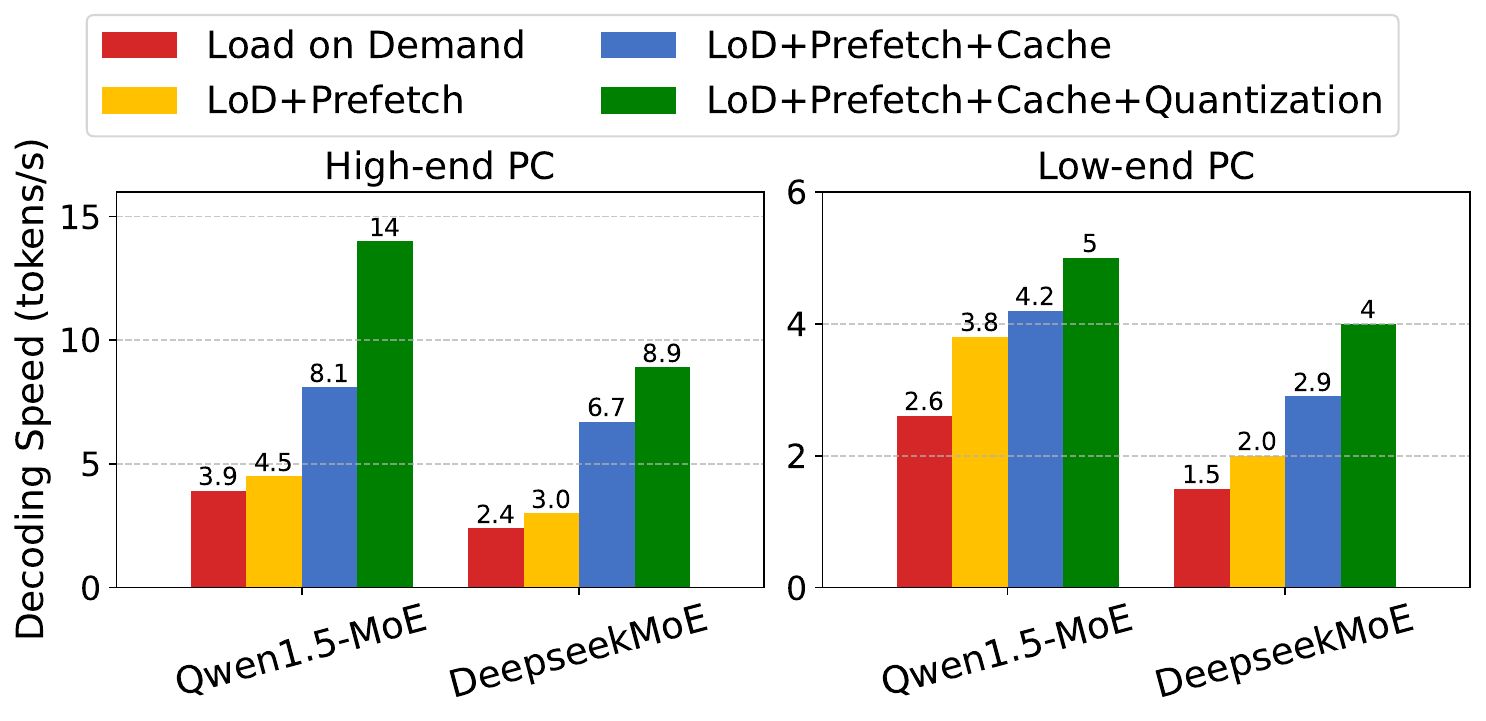}
  \caption{Ablation study of Fate.}
  \label{figure:ablation}
\end{figure}

We conducted a comprehensive breakdown analysis of the benefits brought by each technique in Fate as shown in~\autoref{figure:ablation}. Starting with Load on Demand as a naive baseline, we incrementally incorporated the prefetching strategy (§~\ref{4.3}), the caching strategy (§~\ref{4.4}), and the quantization strategy (§~\ref{4.5}). We then compared their contributions to improving decoding speed.

Specifically, in each scenario, the decoding speed increases progressively with each added component, directly demonstrating the effectiveness of each strategy. First, while the improvement from prefetching may seem modest, \autoref{figure:mem_budget} shows that inaccurate prefetching can lead to negative effects. Therefore, achieving an average speedup of 30\% from prefetching alone confirms the high accuracy of the prefetching strategy, especially on low-end PCs. The caching strategy, by buffering a significant number of experts, effectively reduces I/O overhead, with benefits particularly evident on high-end PCs with larger memory capacity. Building on prefetching, caching achieves up to a 2× decoding speedup. Quantization, in addition to reducing I/O overhead,  allows more experts to be cached and achieves the most significant improvement, delivering up to a 3.7× decoding speedup compared to Load on Demand. Overall, these results highlight the complementary benefits of prefetching, caching, and quantization in improving inference efficiency.

\section{Related Work}

To address the challenges posed by memory bottlenecks, existing efforts to achieve efficient LLM inference in offloading scenarios fall into two main approaches: model compression and offloading systems.

\textbf{Model Compression.} Model compression includes techniques such as knowledge distillation~\cite{agarwal2023gkd, gu2023knowledge, li2023symbolic}, pruning~\cite{frantar2023sparsegpt, ma2023llm, sun2023simple}, quantization~\cite{frantar2022gptq, lin2024awq, xiao2023smoothquant}, low-rank factorization~\cite{xu2023tensorgpt}, etc. DeepSpeedMoE~\cite{rajbhandari2022deepspeed} uses staged knowledge distillation to produce a Mixture-of-Students (MoS), reducing model size by 12.5\% while retaining 99.3\% of the performance of the teacher model. QMoE~\cite{frantar2023qmoe} compresses the trillion-parameter SwitchTransformer-c2048 into a custom format, achieving sub-1-bit parameter representation, allowing such massive models to be deployed on a single GPU. And there is a work~\cite{lu2024not} suggesting that not all experts are equally important, so certain experts can be pruned or skipped during inference. 
Additionally, many compression techniques not specifically designed for MoE models can still provide significant benefits for them. Importantly, Fate is orthogonal to these compression efforts, and each module in Fate is decoupled. For example, Fate allows the seamless integration of faster and more accurate quantization algorithms to further improve MoE inference performance.

\textbf{Offloading Systems.} Offloading systems have been instrumental in enabling efficient LLM inference under resource constraints. For instance, DeepSpeed-Inference~\cite{aminabadi2022deepspeed} pioneered the application of ZeRO series~\cite{ren2021zero, rajbhandari2021zero} offloading techniques to LLM inference, facilitating resource-constrained deployments. FlexGen~\cite{sheng2023flexgen} uses linear programming to optimise computational graphs, significantly improving inference throughput, even achieving over 1 token/s for 175B OPT~\cite{zhang2022opt} on a single GPU. However, these offloading systems are primarily designed for dense models and do not take into account the sparse activation characteristics of MoE models.
To address this gap, Mixtral-offloading~\cite{eliseev2023fast} introduced the use of an LRU cache and quantization to rapidly load a subset of experts, enabling inference for Mixtral-8x7B~\cite{jiang2024mixtral} on consumer-grade hardware. Similarly, MoE-Infinity~\cite{xue2024moe} reduced the latency costs associated with expert offloading through novel activation-aware expert prefetching and caching strategies. Pre-gated MoE~\cite{hwang2024pre} trained a pre-gate to replace the original gating mechanism, enabling early routing decisions. SiDA~\cite{du2024sida}, by training offline hash functions, achieved over a 90\% hash hit rate, significantly reducing expert selection overhead.
However, these methods often suffer from trade-offs such as limited routing accuracy or additional training overhead, which can reduce their practicality. In contrast, Fate overcomes the challenges of expert prediction with a simple yet highly accurate cross-layer prefetching strategy, achieving efficient MoE inference under offloading scenarios.

\section{Conclusion}

This paper presents Fate, a system designed for fast MoE inference in edge scenarios. The key insight of Fate lies in leveraging the high cosine similarity of adjacent gate inputs to achieve accurate cross-layer prefetching. In addition, a shallow-favoring expert caching strategy is designed to improve the expert hit rate to over 99\%. Quantization techniques are also integrated to optimize both I/O and caching.
Experimental results show that Fate effectively accelerates the prefill and decoding speed of MoE models while maintaining inference accuracy. Furthermore, these performance improvements are scalable over varying memory budgets.

\begin{acks}
The work described in this paper was supported by CAAI-MindSpore Open Fund, developed on OpenI Community (CAAIXSJLJJ 2023 MindSpore 01).
\end{acks}

\bibliographystyle{plain}
\balance
\bibliography{sample}

\end{document}